\documentclass[submission,copyright,creativecommons]{eptcs}
\providecommand{\event}{AREA 2023} 

\usepackage{iftex}
\usepackage{graphicx}
\usepackage{latexsym}
\usepackage{cite}
\usepackage{amsmath,amssymb,amsfonts}
\usepackage{algorithmic}
\usepackage{textcomp}
\usepackage{xcolor}
\usepackage{multicol}
\usepackage{multirow}
\usepackage[greek,english]{babel}
\usepackage[utf8]{inputenc}
\usepackage{subfig}
\usepackage{placeins}
\usepackage[linesnumbered, ruled]{algorithm2e} 
\ifpdf
  \usepackage{underscore}         
  \usepackage[T1]{fontenc}        
\else
  \usepackage{breakurl}           
\fi

\title{From Robot Self-Localization to Global-Localization:\\  An RSSI Based Approach.}
\author{Athanasios Lentzas
\institute{Aristotle University\\ Thessaloniki, Greece}
\institute{School of Informatics\\
Aristotle University of Thessaloniki\\
Thessaloniki, Greece}
\email{alentzas@csd.auth.gr}
\and
Dimitris Vrakas
\institute{Aristotle University\\ Thessaloniki, Greece}
\institute{School of Informatics\\
Aristotle University of Thessaloniki\\
Thessaloniki, Greece}
\email{dvrakas@csd.auth.gr}
}
\def\titlerunning{From Robot Self-Localization to Global-Localization: An RSSI Based Approach.}
\def\authorrunning{A. Lentzas \& D. Vrakas}
\begin{document}
\maketitle

\begin{abstract}
Localization is a crucial task for autonomous mobile robots in order to successfully move to goal locations in their environment. Usually, this is done in a robot-centric manner, where the robot maintains a map with its body in the center. In swarm robotics applications, where a group of robots needs to coordinate in order to achieve their common goals, robot-centric localization will not suffice as each member of the swarm has its own frame of reference. One way to deal with this problem is to create, maintain and share a common map (global coordinate system), among the members of the swarm. This paper presents an approach to global localization for a group of robots in unknown, GPS and landmark free environments. The main idea relies on members of the swarm staying still and acting as beacons, emitting electromagnetic signals. These stationary robots form a global frame of reference and the rest of the group localize themselves in it using the Received Signal Strength Indicator (RSSI). The proposed method is evaluated, and the results obtained from the experiments are promising.
\end{abstract}

\section{Introduction}
Robotic Swarms is a topic with significant interest, especially with the increased use of large groups of unmanned air or ground vehicles \cite{NEDJAH2019100565}. Self localization and navigation is a crucial task for autonomous robots \cite{75902}, both on indoor and outdoor navigation. Depending on the application of the robotic swarm, the robots may need the ability to discover their location. 

Non robotic applications can also take advantage of indoor localization techniques. In case of emergency evacuation, especially on crowded buildings like malls, airports etc., it is beneficial to know the location of people inside the building as this can help with faster and safer evacuation. In order to achieve that, specific devices or smartphones can be used that take advantage of the proposed localization scheme. 

Localization is the ability of a robot to know its position and orientation. The location could be relative to other robots or absolute on a common coordinate system. While operating outdoors, a robot can rely on GPS to localize itself. The aforementioned sensor can not be used on indoor environments. Simultaneous localization and mapping (SLAM) algorithms are extensively used \cite{Bresson2017}, although when small and relatively simple robots are needed, their high complexity and computational cost is a major drawback. Several techniques exploiting beacons and landmarks for localization have been proposed in the literature as well as techniques using  inertial navigation systems, magnetic, sound and optic based navigation \cite{Obeidat2021}.

Inertial localization is also a common approach. Based on odometers, accelerometers and gyroscopes, one can determine the orientation and direction of the robot \cite{10.1007/978-3-319-94274-2_13}. Although the results provided are robust against environmental changes,  this technique is prone to error accumulation \cite{Petritoli}. Kalman filters have been expensively used in order to improve accuracy \cite{s20061578}.

Radio frequency localization is the approach usually preferred. Beacons can cover a wide area, radio waves can penetrate most materials while the installation cost is relatively low \cite{liu_2014}. Received Signal Strength Indicator (RSSI) is mainly exploited for robot localization \cite{9165300,10.1145/2543581.2543592,5073315}. Time of arrival and time difference of arrival of two signals that are known to have different propagation is also a common approach for localization.

Localization based on RSSI is proposed in the Ladybug algorithm \cite{9212115}. A robot equipped with sensors able to measure the strength of the received electromagnetic signal, is able to identify the location of the source of the signal and navigate to it. The source could be either a beacon or another robot. The LadyBug Algorithm was effective and had numerous benefits compared to similar approaches, such as I-Bug \cite{5152728}. RSSI could be an efficient solution when deploying a robotic swarm on GPS denying environment. Our motivation was to propose an approach where a robotic swarm could be able to extract the location of each robot in the same coordinate system using local sensing only, allowing the LadyBug algorithm to be implemented on the swarm. Sharing a common coordinate system is crucial for tasks such as self-assembly. 

The benefits of localizing the swarm in a common coordinate system are presented in \cite{doi:10.1126/science.1254295}. The ability of the swarm to self-assemble on a specific shape was realized by placing four robots on a specific layout, marking the coordinate system and using trillateration. While the proposed solution is efficient, it's main drawback is the limitation to two dimensions and the requirement of hand placing the four initial robots. Our incentive was to propose an approach where the robots will use local sensing but will have the ability to localize in a global system without the aforementioned limitations. 

Our approach would have numerous benefits such as: a) the swarm will share a common localization scheme, sharing the same map. b) The localization scheme is based on three members of the swarm equipped with a beacon instead of four. c) No manual placement is required. The beacons could be anywhere on the operation area. d) Our approach is able to localize in three dimensions. e) The proposed solution is robust as, in case of a possible failure of a beacon, another robot from the swarm can replace it. 

\section{Related Work}
Radio frequency identification technology (RFID) is used in \cite{TESORIERO2010894} to localize robots navigating indoors. The passive RFID tags installed, divided the area into a grid. As a robot, equipped with an RFID reader, explores the area, the reader reads the tag. The position is then estimated by correlating the ID of the tag with a map containing the localization information of the tags. Despite the high accuracy, the need for a map containing the ID and location of the tags limits the practical applications.  

Pseudolites (i.e. pseudo satellites) have been proposed for indoor localization \cite{WAN20111446}. The main idea is to receive the GPS signal and transmit it indoors using signal repeaters \cite{s150510074}. The major drawback of this approach is the increased cost of the network installation. Additionally, in case the main signal receiver fails the whole network is unusable.

A localization method based on RSSI of heterogeneous sources (i.e. WLAN, GSM etc) is presented in \cite{STELLA20146738}. By analyzing the fingerprint and strength of the received signal, the robot is able to localize itself by comparing it with a fingerprint map. While exploiting already existing infrastructures, negating the need for a network deployment, the main disadvantage is the need for a fingerprint map covering all the area of interest. RFID sensors can also be used for indoor localization. In \cite{9715113}, authors presented a simulator that allows modeling environments and testing the deployment of RFID solutions. Tags and antennas are placed and RSSI is exploited to find the location of each tag.   

A distributed localization method based on swarm intelligence algorithms is presented in \cite{DESA2016322}. Particle swarm optimization and an approach based on backtracking search algorithm is proposed. The proposed method operates in three stages. The first stage allows a robot to estimate the distance from a reference node using the Sum-Dist algorithm \cite{LANGENDOEN2003499}. The second stage estimates the position of the robot using the min-max technique\cite{10.1145/570738.570755,LANGENDOEN2003499}. Finally the third stage the accuracy improves by re-evaluating the position based on the position of the neighbors.

\section{Problem Statement}
  The localization of the robot is based on RSSI. Given an electromagnetic signal, the coordinates of the source in 3D space can be calculated in the robocentric system and the in the global coordinate system. In order to achieve our goal, three beacons are employed, creating a 3D common coordinate system for all robots.  In our work, the following assumptions were made: a) All robots are equipped with sensors capable of reading the RSSI of a received signal. b) Two robots act as anchors. c) A beacon/robot is placed at the center of the coordinate system.

\begin{figure}[htbp]
\centerline{\includegraphics[width=0.15\textwidth]{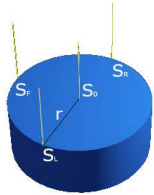}}
\caption{The position of the four sensors on the robot.}
\label{robot}
\end{figure}

The main localization scheme employs three receivers and one transceiver. The transceiver allows the robot to act as a beacon or exchange information with the rest of the swarm. The receivers are placed on the front, right and left of a robot with radius \textit{r} and the transceiver on the center, as seen in Fig. \ref{robot}. Each sensor is able to  identify both  the source of the signal and the signal strength. The signal strength (\textit{S}) is inversely proportional to the square of the distance (\textit{d}) between the sensor and the source \eqref{lsensor}. Using \eqref{lsensor} the distance between the source and each sensor can be calculated \eqref{lsensor}.

\begin{align}
 S&=\frac{1}{d^2}          &  d&=\sqrt{\frac{1}{S}} \label{lsensor}  
\end{align}

 The coordinates of each sensor, in the robot-centric system, can be seen on Table \ref{coordinates}. Let $(x_s,y_s,z_s)$ be the coordinates of the signal source in the robot-centric system. Using \eqref{lsensor} the distance between each sensor (center, front, left, right) is calculated.
 \begin{table}[]
    \centering
    \begin{tabular}{c|c|c}
        \hline
        \textbf{Sensor} & \textbf{Robocentric Coordinates (x,y)} & \textbf{Enchanced Localization Coordinates}\\
        \hline
        $S_0$&(0,0,0)&(0,0)\\
        $S_L$&(0,r,0)&(0,-r)\\
        $S_R$&(0,-r,0)&(0,r)\\
        $S_F$&(r,0,0)&(r,0)\\
        $S_B$& - &(-r,0)\\
        \hline
    \end{tabular}
    \caption{Position of the sensors}
    \label{coordinates}
\end{table}

\begin{align} \label{distances}
    d_0 ^ 2&=x_s ^ 2 + y_s ^2 + z_s ^ 2 &
    d_f ^ 2&=x_s ^ 2 +2rx +r^2 + y_s ^2 + z_s^2 \nonumber \\
    d_l ^ 2&=x_s ^ 2 + y_s ^2 + z_s^2 +2ry+r^2 &
    d_r ^ 2&=x_s ^ 2 + y_s ^2 + z_s^2 -2ry+r^2 
\end{align} 

Solving the equation system \eqref{distances}, the coordinates of the source, $x_s$ and $y_s$ respectively, can be calculated. By replacing the values of $x_s$ and $y_s$ on \eqref{distances} the $z_s$ coordinate can be calculated. Similarly (and knowing the coordinates of the source) the coordinates of the beacon $(x_b,y_b, z_b)$ and the anchor robots $(x_{a1},y_{a1},z_{a1})$, $(x_{a2},y_{a2},z_{a2})$ can be calculated.

\subsection{Enhanced Localization}
In order to further enhance the localization procedure, redundant sensors were added. Our implementation uses five sensors. All the receivers are placed on the perimeter of the robot  thus creating four sensor groups (triangle formed by the transceiver in the center and two receivers on the perimeter). The updated robot-centric coordinates can be seen on Table \ref{coordinates}. The estimates of the four groups are averaged, resulting in more accurate estimation of the source's position.

\subsection{Global Localization}
As already described in the previous section, the robot is able to calculate the coordinates of the three beacons in the robocentric system. Knowing their relative positions, the global position of the robot can be calculated. Let \textit{A} be the center of the global coordinate system and \textit{B}, \textit{C} the two beacons that define axis X and Y respectively. 

Using vectors $\Vec{AB}$ and $\Vec{AC}$ we define three new vectors using the outer products, as seen in \eqref{oprod}, and normalize them.

\begin{align}\label{oprod}
    \Vec{z}&=\Vec{AB} \otimes \Vec{AC}  & \Vec{y}&=\Vec{z} \otimes \Vec{AB}  &\Vec{x}&=\Vec{y} \otimes \Vec{z} 
\end{align} 
 
Let $\Vec{r}$ be the vector between the center of the robot \textit{R} and \textit{A}. We can now calculate the coordinates of the robot in the global system defined by $\Vec{x}$, $\Vec{y}$ and $\Vec{z}$. The coordinates of the robot, $(x_r, y_r, z_r)$, are the dot product of $\Vec{r}$ and  $\Vec{x}$, $\Vec{y}$, $\Vec{z}$ respectively. 

\section{Experimental Results}
In order to evaluate the performance of our localization approach we implemented the algorithm in the Webots Open Source Robot simulator \cite{Webots}. A swarm of four identical robots was used, where three of them were serving as beacons. A generic round shaped robot with 20cm radius was employed, equipped with four radio receivers capable of identifying the signal strength on its perimeter and a radio transceiver on the center. 

The three robots acting as beacons were constantly emitting a radio signal with a unique ID, allowing the robot to identify the source of the signal. In order to further increase the accuracy, each beacon was transmitting on a different channel reducing the generated noise. For each source the RSSI of 100 packets was averaged reducing the noise generated error. The robot was cycling through the three predefined channels allowing it to identify and locate the robocentric position of each beacon. Lastly the global coordinates were calculated and reported. 

During the experiment, the robot was static while calculating it's position. After a successfully calculation it was moving randomly and the new position was calculated and reported. This process was repeated 10 times. The error (i.e. euclidean distance between the real and calculated global location) for each location was averaged and used as a metric. In a noiseless environment, the calculated position was the same with the real one, no matter the distance from the beacons or the orientation of the robot. 

\begin{figure}[htbp]
\centering
\subfloat[Error per noise level]{\includegraphics[width=0.4\textwidth]{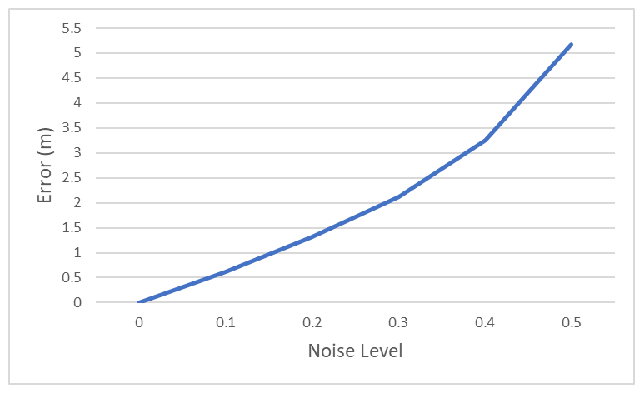}\label{noise}}
\hfill
\subfloat[Error per distance from the beacons]{\includegraphics[width=0.4\textwidth]{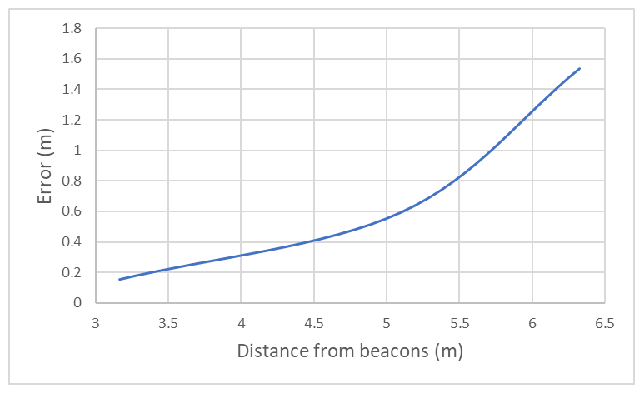}\label{noisedist}}
\caption{Localization error.}
\label{error}
\end{figure}

In order to evaluate our approach in more real-like conditions, noise was gradually introduced into the experiment. In total the process described above was repeated 5 times with 5 different noise levels: 10\%, 20\%, 30\%, 40\% and 50\%. Noise strength is the standard deviation of the Gaussian noise added to the signal strength. The rest of the conditions were controlled (i.e. the random seed) making sure the experiment would be the same, providing comparable results. As seen in Fig. \ref{noise}, for 10\% noise the error was ~0.6m while for 20\% noise the calculated position was ~1.4m off. Further increasing the noise had a bigger impact. 

As noise is related to the distance of the source, an experiment was performed where the distance between the robot and beacons was gradually increased. The three beacons were hand placed forming an isosceles triangle. This placement guaranteed that the distance between the robot and two beacons would be the same. The side of the triangle formed was 4m and the noise level was set to 10\%. The experiment was repeated 10 times and the results were averaged. As seen in Fig. \ref{noisedist} the error was ~0.1m when the robot was closer to the source. The best results were observed when the robot was between 3m and 3.5m from the beacons. Increasing the distance between the robot and source results in increased noise level, making it harder to correctly identify the position of the robot. It is worth mentioning that for distances up to almost 6m the calculated position error was less than 1m. 

\begin{figure}[htbp]
\centerline{\includegraphics[width=0.4\textwidth]{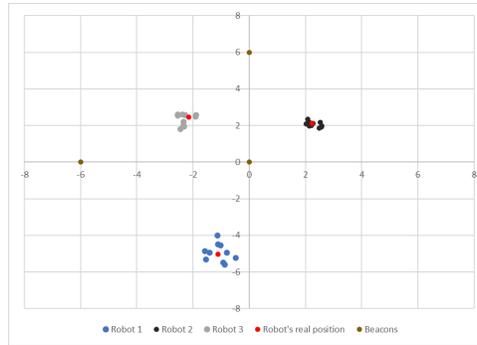}}
\caption{Calculated position of each robot.}
\label{swarm}
\end{figure}

An experiment was also performed using a small swarm of three robots. The first robot was placed away from the beacons, the second close to the two of them while the third close to the three beacon robots. The robots reported their global position 10 times with a time delay added between each calculation. The noise level was set to 10\%. Further noise was introduced, as each robot was emitting a random signal, simulating an environment where the robots communicate with each other. In Fig. \ref{swarm} the position of each robot can be seen. The results show that even with the extra noise (i.e. robots communicating with each other), the localization process is not heavily affected. It is worth mentioning that the distance between the robot and the beacons has an impact on the calculations (as already discussed). 

\section{Conclusion \& Future Work}
In this paper, a novel global localization method was presented. Our approach exploits RSSI from a signal emitted by three beacons, to localize a robot on the coordinate system formed by those beacons. 

Our approach was evaluated in a noiseless environment as well as in an environment with different noise levels. While for lower noise levels the results were accepted, for increased noise levels the error was bigger. This requires further experiments as noise levels in the real world are affected by many factors, such as the materials of the obstacles, reflectance etc. 

Additionally, the impact of the distance between the robot and the source of the signal was investigated. Placing the robot close to the beacons negatively affects the ability to localize itself. Increasing the distance up to a certain threshold  has low impact. Further increasing the distance produced increased error. 

Although a small swarm was simulated, allowing the observation of how multiple robots communicating will impact the performance, the robots were not exchanging any location related information. In the near future our focus will be on investigating ways that will allow the robots to correlate their individual localization information and create a more accurate global map. 

Furthermore, the current experiments were performed in a simulated environment. Even with noise added it is not as close as the real world. Experiments will be performed on a real robot. That evaluation is crucial as real world noise can not be reproduced easily (i.e. reflection of the signal, non-uniform noise etc.). 

Another important part of our future work is to integrate the localization presented with the LadyBug algorithm \cite{5152728}. As LadyBug provides robust results, the proposed method will allow its application to robotic swarms.

\section{Acknowledgements} This research was carried out as part of the project «Beacon: An Intelligent Dynamic Signage System for Emergency Situations» (Project code: KMP6-0094837) under the framework of the Action «Investment Plans of Innovation» of the Operational Program «Central Macedonia 2014 2020», which is co-funded by the European Regional Development Fund and Greece.

\nocite{*}
\bibliographystyle{eptcs}
\bibliography{generic}

\begin{thebibliography}{10}
\providecommand{\bibitemdeclare}[2]{}
\providecommand{\surnamestart}{}
\providecommand{\surnameend}{}
\providecommand{\urlprefix}{Available at }
\providecommand{\url}[1]{\texttt{#1}}
\providecommand{\href}[2]{\texttt{#2}}
\providecommand{\urlalt}[2]{\href{#1}{#2}}
\providecommand{\doi}[1]{doi:\urlalt{https://doi.org/#1}{#1}}
\providecommand{\eprint}[1]{arXiv:\urlalt{https://arxiv.org/abs/#1}{#1}}
\providecommand{\bibinfo}[2]{#2}

\bibitemdeclare{inproceedings}{5073315}
\bibitem{5073315}
\bibinfo{author}{Paolo \surnamestart Barsocchi\surnameend},
  \bibinfo{author}{Stefano \surnamestart Lenzi\surnameend},
  \bibinfo{author}{Stefano \surnamestart Chessa\surnameend} \&
  \bibinfo{author}{Gaetano \surnamestart Giunta\surnameend}
  (\bibinfo{year}{2009}): \emph{\bibinfo{title}{A Novel Approach to Indoor RSSI
  Localization by Automatic Calibration of the Wireless Propagation Model}}.
\newblock In: {\slshape \bibinfo{booktitle}{VTC Spring 2009 - IEEE 69th
  Vehicular Technology Conference}}, pp. \bibinfo{pages}{1--5},
  \doi{10.1109/VETECS.2009.5073315}.

\bibitemdeclare{article}{Bresson2017}
\bibitem{Bresson2017}
\bibinfo{author}{G.~\surnamestart {Bresson}\surnameend},
  \bibinfo{author}{Z.~\surnamestart {Alsayed}\surnameend},
  \bibinfo{author}{L.~\surnamestart {Yu}\surnameend} \&
  \bibinfo{author}{S.~\surnamestart {Glaser}\surnameend}
  (\bibinfo{year}{2017}): \emph{\bibinfo{title}{Simultaneous Localization and
  Mapping: A Survey of Current Trends in Autonomous Driving}}.
\newblock {\slshape \bibinfo{journal}{IEEE Transactions on Intelligent
  Vehicles}} \bibinfo{volume}{2}(\bibinfo{number}{3}), pp.
  \bibinfo{pages}{194--220}, \doi{10.1109/TIV.2017.2749181}.

\bibitemdeclare{article}{75902}
\bibitem{75902}
\bibinfo{author}{I.J. \surnamestart Cox\surnameend} (\bibinfo{year}{1991}):
  \emph{\bibinfo{title}{Blanche-an experiment in guidance and navigation of an
  autonomous robot vehicle}}.
\newblock {\slshape \bibinfo{journal}{IEEE Transactions on Robotics and
  Automation}} \bibinfo{volume}{7}(\bibinfo{number}{2}), pp.
  \bibinfo{pages}{193--204}, \doi{10.1109/70.75902}.

\bibitemdeclare{article}{DESA2016322}
\bibitem{DESA2016322}
\bibinfo{author}{Alan~Oliveira \surnamestart {de Sa}\surnameend},
  \bibinfo{author}{Nadia \surnamestart Nedjah\surnameend} \&
  \bibinfo{author}{Luiza \surnamestart {de Macedo Mourelle}\surnameend}
  (\bibinfo{year}{2016}): \emph{\bibinfo{title}{Distributed efficient
  localization in swarm robotic systems using swarm intelligence algorithms}}.
\newblock {\slshape \bibinfo{journal}{Neurocomputing}} \bibinfo{volume}{172},
  pp. \bibinfo{pages}{322--336}, \doi{10.1016/j.neucom.2015.03.099}.
\newblock
  \urlprefix\url{https://www.sciencedirect.com/science/article/pii/S0925231215010498}.

\bibitemdeclare{inproceedings}{6402492}
\bibitem{6402492}
\bibinfo{author}{Qian \surnamestart Dong\surnameend} \&
  \bibinfo{author}{Waltenegus \surnamestart Dargie\surnameend}
  (\bibinfo{year}{2012}): \emph{\bibinfo{title}{Evaluation of the reliability
  of RSSI for indoor localization}}.
\newblock In: {\slshape \bibinfo{booktitle}{2012 International Conference on
  Wireless Communications in Underground and Confined Areas}}, pp.
  \bibinfo{pages}{1--6}, \doi{10.1109/ICWCUCA.2012.6402492}.

\bibitemdeclare{article}{9715113}
\bibitem{9715113}
\bibinfo{author}{Salvatore \surnamestart D’Avella\surnameend},
  \bibinfo{author}{Matteo \surnamestart Unetti\surnameend} \&
  \bibinfo{author}{Paolo \surnamestart Tripicchio\surnameend}
  (\bibinfo{year}{2022}): \emph{\bibinfo{title}{RFID Gazebo-Based Simulator
  With RSSI and Phase Signals for UHF Tags Localization and Tracking}}.
\newblock {\slshape \bibinfo{journal}{IEEE Access}} \bibinfo{volume}{10}, pp.
  \bibinfo{pages}{22150--22160}, \doi{10.1109/ACCESS.2022.3152199}.

\bibitemdeclare{inproceedings}{10.1007/978-3-319-94274-2_13}
\bibitem{10.1007/978-3-319-94274-2_13}
\bibinfo{author}{Giovanni \surnamestart Fusco\surnameend} \&
  \bibinfo{author}{James~M. \surnamestart Coughlan\surnameend}
  (\bibinfo{year}{2018}): \emph{\bibinfo{title}{Indoor Localization Using
  Computer Vision and Visual-Inertial Odometry}}.
\newblock In \bibinfo{editor}{Klaus \surnamestart Miesenberger\surnameend} \&
  \bibinfo{editor}{Georgios \surnamestart Kouroupetroglou\surnameend}, editors:
  {\slshape \bibinfo{booktitle}{Computers Helping People with Special Needs}},
  \bibinfo{publisher}{Springer International Publishing},
  \bibinfo{address}{Cham}, pp. \bibinfo{pages}{86--93},
  \doi{10.1007/978-3-319-94274-2_13}.

\bibitemdeclare{article}{s20061578}
\bibitem{s20061578}
\bibinfo{author}{Guanghui \surnamestart Hu\surnameend}, \bibinfo{author}{Weizhi
  \surnamestart Zhang\surnameend}, \bibinfo{author}{Hong \surnamestart
  Wan\surnameend} \& \bibinfo{author}{Xinxin \surnamestart Li\surnameend}
  (\bibinfo{year}{2020}): \emph{\bibinfo{title}{Improving the Heading Accuracy
  in Indoor Pedestrian Navigation Based on a Decision Tree and Kalman Filter}}.
\newblock {\slshape \bibinfo{journal}{Sensors}}
  \bibinfo{volume}{20}(\bibinfo{number}{6}), \doi{10.3390/s20061578}.
\newblock \urlprefix\url{https://www.mdpi.com/1424-8220/20/6/1578}.

\bibitemdeclare{article}{LANGENDOEN2003499}
\bibitem{LANGENDOEN2003499}
\bibinfo{author}{Koen \surnamestart Langendoen\surnameend} \&
  \bibinfo{author}{Niels \surnamestart Reijers\surnameend}
  (\bibinfo{year}{2003}): \emph{\bibinfo{title}{Distributed localization in
  wireless sensor networks: a quantitative comparison}}.
\newblock {\slshape \bibinfo{journal}{Computer Networks}}
  \bibinfo{volume}{43}(\bibinfo{number}{4}), pp. \bibinfo{pages}{499--518},
  \doi{10.1016/S1389-1286(03)00356-6}.
\newblock
  \urlprefix\url{https://www.sciencedirect.com/science/article/pii/S1389128603003566}.
\newblock \bibinfo{note}{Wireless Sensor Networks}.

\bibitemdeclare{inproceedings}{9212115}
\bibitem{9212115}
\bibinfo{author}{Athanasios \surnamestart Lentzas\surnameend} \&
  \bibinfo{author}{Dimitris \surnamestart Vrakas\surnameend}
  (\bibinfo{year}{2020}): \emph{\bibinfo{title}{LadyBug. An Intensity based
  Localization Bug Algorithm}}.
\newblock In: {\slshape \bibinfo{booktitle}{2020 25th IEEE International
  Conference on Emerging Technologies and Factory Automation (ETFA)}},
  \bibinfo{volume}{1}, pp. \bibinfo{pages}{682--689},
  \doi{10.1109/ETFA46521.2020.9212115}.

\bibitemdeclare{article}{liu_2014}
\bibitem{liu_2014}
\bibinfo{author}{Junjie \surnamestart Liu\surnameend} (\bibinfo{year}{2014}):
  \emph{\bibinfo{title}{Survey of Wireless Based Indoor
  LocalizationTechnologies}}.
\newblock {\slshape \bibinfo{journal}{Department of Science and Engineering
  Washington University}}.

\bibitemdeclare{article}{Webots04}
\bibitem{Webots04}
\bibinfo{author}{O.~\surnamestart Michel\surnameend} (\bibinfo{year}{2004}):
  \emph{\bibinfo{title}{Webots: Professional Mobile Robot Simulation}}.
\newblock {\slshape \bibinfo{journal}{Journal of Advanced Robotics Systems}}
  \bibinfo{volume}{1}(\bibinfo{number}{1}), pp. \bibinfo{pages}{39--42}.
\newblock \urlprefix\url{http://www.ars-journal.com/International-Journal-of-
  Advanced-Robotic-Systems/Volume-1/39-42.pdf}.

\bibitemdeclare{article}{NEDJAH2019100565}
\bibitem{NEDJAH2019100565}
\bibinfo{author}{Nadia \surnamestart Nedjah\surnameend} \&
  \bibinfo{author}{Luneque~Silva \surnamestart Junior\surnameend}
  (\bibinfo{year}{2019}): \emph{\bibinfo{title}{Review of methodologies and
  tasks in swarm robotics towards standardization}}.
\newblock {\slshape \bibinfo{journal}{Swarm and Evolutionary Computation}}
  \bibinfo{volume}{50}, p. \bibinfo{pages}{100565},
  \doi{10.1016/j.swevo.2019.100565}.
\newblock
  \urlprefix\url{https://www.sciencedirect.com/science/article/pii/S2210650217308398}.

\bibitemdeclare{article}{Obeidat2021}
\bibitem{Obeidat2021}
\bibinfo{author}{Huthaifa \surnamestart Obeidat\surnameend},
  \bibinfo{author}{Wafa \surnamestart Shuaieb\surnameend},
  \bibinfo{author}{Omar \surnamestart Obeidat\surnameend} \&
  \bibinfo{author}{Raed \surnamestart Abd-Alhameed\surnameend}
  (\bibinfo{year}{2021}): \emph{\bibinfo{title}{A Review of Indoor Localization
  Techniques and Wireless Technologies}}.
\newblock {\slshape \bibinfo{journal}{Wireless Personal Communications}},
  \doi{10.1007/s11277-021-08209-5}.

\bibitemdeclare{inproceedings}{Petritoli}
\bibitem{Petritoli}
\bibinfo{author}{Enrico \surnamestart Petritoli\surnameend},
  \bibinfo{author}{Fabio \surnamestart Leccese\surnameend} \&
  \bibinfo{author}{Mariagrazia \surnamestart Leccisi\surnameend}
  (\bibinfo{year}{2019}): \emph{\bibinfo{title}{Inertial Navigation Systems for
  UAV: Uncertainty and Error Measurements}}.
\newblock In: {\slshape \bibinfo{booktitle}{2019 IEEE 5th International
  Workshop on Metrology for AeroSpace (MetroAeroSpace)}}, pp.
  \bibinfo{pages}{1--5}, \doi{10.1109/MetroAeroSpace.2019.8869618}.

\bibitemdeclare{inproceedings}{Rothermich}
\bibitem{Rothermich}
\bibinfo{author}{Joseph~A. \surnamestart Rothermich\surnameend},
  \bibinfo{author}{M.~{\.{I}}hsan \surnamestart Ecemi{\c{s}}\surnameend} \&
  \bibinfo{author}{Paolo \surnamestart Gaudiano\surnameend}
  (\bibinfo{year}{2005}): \emph{\bibinfo{title}{Distributed Localization and
  Mapping with a Robotic Swarm}}.
\newblock In \bibinfo{editor}{Erol \surnamestart {\c{S}}ahin\surnameend} \&
  \bibinfo{editor}{William~M. \surnamestart Spears\surnameend}, editors:
  {\slshape \bibinfo{booktitle}{Swarm Robotics}}, \bibinfo{publisher}{Springer
  Berlin Heidelberg}, \bibinfo{address}{Berlin, Heidelberg}, pp.
  \bibinfo{pages}{58--69}, \doi{10.1007/978-3-540-30552-1_6}.

\bibitemdeclare{inbook}{Roumeliotis2000}
\bibitem{Roumeliotis2000}
\bibinfo{author}{Stergios~I. \surnamestart Roumeliotis\surnameend} \&
  \bibinfo{author}{George~A. \surnamestart Bekey\surnameend}
  (\bibinfo{year}{2000}): \emph{\bibinfo{title}{Distributed Multi-Robot
  Localization}}, pp. \bibinfo{pages}{179--188}.
\newblock \bibinfo{publisher}{Springer Japan}, \bibinfo{address}{Tokyo},
  \doi{10.1007/978-4-431-67919-6_17}.

\bibitemdeclare{article}{doi:10.1126/science.1254295}
\bibitem{doi:10.1126/science.1254295}
\bibinfo{author}{Michael \surnamestart Rubenstein\surnameend},
  \bibinfo{author}{Alejandro \surnamestart Cornejo\surnameend} \&
  \bibinfo{author}{Radhika \surnamestart Nagpal\surnameend}
  (\bibinfo{year}{2014}): \emph{\bibinfo{title}{Programmable self-assembly in a
  thousand-robot swarm}}.
\newblock {\slshape \bibinfo{journal}{Science}}
  \bibinfo{volume}{345}(\bibinfo{number}{6198}), pp. \bibinfo{pages}{795--799},
  \doi{10.1126/science.1254295}.

\bibitemdeclare{inproceedings}{9165300}
\bibitem{9165300}
\bibinfo{author}{Jirapat \surnamestart Sangthong\surnameend},
  \bibinfo{author}{Jutamas \surnamestart Thongkam\surnameend} \&
  \bibinfo{author}{Sathapom \surnamestart Promwong\surnameend}
  (\bibinfo{year}{2020}): \emph{\bibinfo{title}{Indoor Wireless Sensor Network
  Localization Using RSSI Based Weighting Algorithm Method}}.
\newblock In: {\slshape \bibinfo{booktitle}{2020 6th International Conference
  on Engineering, Applied Sciences and Technology (ICEAST)}}, pp.
  \bibinfo{pages}{1--4}, \doi{10.1109/ICEAST50382.2020.9165300}.

\bibitemdeclare{inproceedings}{10.1145/570738.570755}
\bibitem{10.1145/570738.570755}
\bibinfo{author}{Andreas \surnamestart Savvides\surnameend},
  \bibinfo{author}{Heemin \surnamestart Park\surnameend} \&
  \bibinfo{author}{Mani~B. \surnamestart Srivastava\surnameend}
  (\bibinfo{year}{2002}): \emph{\bibinfo{title}{The Bits and Flops of the N-Hop
  Multilateration Primitive for Node Localization Problems}}.
\newblock In: {\slshape \bibinfo{booktitle}{Proceedings of the 1st ACM
  International Workshop on Wireless Sensor Networks and Applications}},
  \bibinfo{series}{WSNA '02}, \bibinfo{publisher}{Association for Computing
  Machinery}, \bibinfo{address}{New York, NY, USA}, p.
  \bibinfo{pages}{112–121}, \doi{10.1145/570738.570755}.

\bibitemdeclare{article}{STELLA20146738}
\bibitem{STELLA20146738}
\bibinfo{author}{M.~\surnamestart Stella\surnameend},
  \bibinfo{author}{M.~\surnamestart Russo\surnameend} \&
  \bibinfo{author}{D.~\surnamestart Begušić\surnameend}
  (\bibinfo{year}{2014}): \emph{\bibinfo{title}{Fingerprinting based
  localization in heterogeneous wireless networks}}.
\newblock {\slshape \bibinfo{journal}{Expert Systems with Applications}}
  \bibinfo{volume}{41}(\bibinfo{number}{15}), pp. \bibinfo{pages}{6738--6747},
  \doi{10.1016/j.eswa.2014.05.016}.
\newblock
  \urlprefix\url{https://www.sciencedirect.com/science/article/pii/S0957417414002966}.

\bibitemdeclare{inproceedings}{5152728}
\bibitem{5152728}
\bibinfo{author}{Kamilah \surnamestart Taylor\surnameend} \&
  \bibinfo{author}{Steven~M. \surnamestart LaValle\surnameend}
  (\bibinfo{year}{2009}): \emph{\bibinfo{title}{I-Bug: An intensity-based bug
  algorithm}}.
\newblock In: {\slshape \bibinfo{booktitle}{2009 IEEE International Conference
  on Robotics and Automation}}, pp. \bibinfo{pages}{3981--3986},
  \doi{10.1109/ROBOT.2009.5152728}.

\bibitemdeclare{article}{TESORIERO2010894}
\bibitem{TESORIERO2010894}
\bibinfo{author}{R.~\surnamestart Tesoriero\surnameend},
  \bibinfo{author}{R.~\surnamestart Tebar\surnameend}, \bibinfo{author}{J.A.
  \surnamestart Gallud\surnameend}, \bibinfo{author}{M.D. \surnamestart
  Lozano\surnameend} \& \bibinfo{author}{V.M.R. \surnamestart
  Penichet\surnameend} (\bibinfo{year}{2010}): \emph{\bibinfo{title}{Improving
  location awareness in indoor spaces using RFID technology}}.
\newblock {\slshape \bibinfo{journal}{Expert Systems with Applications}}
  \bibinfo{volume}{37}(\bibinfo{number}{1}), pp. \bibinfo{pages}{894--898},
  \doi{10.1016/j.eswa.2009.05.062}.
\newblock
  \urlprefix\url{https://www.sciencedirect.com/science/article/pii/S095741740900503X}.

\bibitemdeclare{article}{WAN20111446}
\bibitem{WAN20111446}
\bibinfo{author}{Xiaoguang \surnamestart Wan\surnameend} \&
  \bibinfo{author}{Xingqun \surnamestart Zhan\surnameend}
  (\bibinfo{year}{2011}): \emph{\bibinfo{title}{The Research of Indoor
  Navigation System using Pseudolites}}.
\newblock {\slshape \bibinfo{journal}{Procedia Engineering}}
  \bibinfo{volume}{15}, pp. \bibinfo{pages}{1446--1450},
  \doi{10.1016/j.proeng.2011.08.268}.
\newblock
  \urlprefix\url{https://www.sciencedirect.com/science/article/pii/S1877705811017693}.
\newblock \bibinfo{note}{CEIS 2011}.

\bibitemdeclare{misc}{Webots}
\bibitem{Webots}
\bibinfo{author}{\surnamestart Webots\surnameend}:
  \emph{\bibinfo{title}{http://www.cyberbotics.com}}.
\newblock \urlprefix\url{http://www.cyberbotics.com}.
\newblock \bibinfo{note}{Commercial Mobile Robot Simulation Software}.

\bibitemdeclare{article}{s150510074}
\bibitem{s150510074}
\bibinfo{author}{Rui \surnamestart Xu\surnameend},
  \bibinfo{author}{Wu~\surnamestart Chen\surnameend}, \bibinfo{author}{Ying
  \surnamestart Xu\surnameend} \& \bibinfo{author}{Shengyue \surnamestart
  Ji\surnameend} (\bibinfo{year}{2015}): \emph{\bibinfo{title}{A New Indoor
  Positioning System Architecture Using GPS Signals}}.
\newblock {\slshape \bibinfo{journal}{Sensors}}
  \bibinfo{volume}{15}(\bibinfo{number}{5}), pp. \bibinfo{pages}{10074--10087},
  \doi{10.3390/s150510074}.
\newblock \urlprefix\url{https://www.mdpi.com/1424-8220/15/5/10074}.

\bibitemdeclare{article}{10.1145/2543581.2543592}
\bibitem{10.1145/2543581.2543592}
\bibinfo{author}{Zheng \surnamestart Yang\surnameend}, \bibinfo{author}{Zimu
  \surnamestart Zhou\surnameend} \& \bibinfo{author}{Yunhao \surnamestart
  Liu\surnameend} (\bibinfo{year}{2013}): \emph{\bibinfo{title}{From RSSI to
  CSI: Indoor Localization via Channel Response}}.
\newblock {\slshape \bibinfo{journal}{ACM Comput. Surv.}}
  \bibinfo{volume}{46}(\bibinfo{number}{2}), \doi{10.1145/2543581.2543592}.

\end{thebibliography}
\end{document}